\documentclass[utf8]{FrontiersinHarvard} 
\usepackage{url,hyperref,microtype,subcaption}
\usepackage[onehalfspacing]{setspace}

\usepackage{url,hyperref,microtype}

\usepackage[onehalfspacing]{setspace}
\usepackage{booktabs}
\usepackage{amsmath,amssymb,amsfonts}
\usepackage{algorithmic}
\usepackage{graphicx}
\usepackage{textcomp}
\usepackage{xcolor}
\usepackage[utf8]{inputenc}
\usepackage{authblk}
\usepackage{float}
\usepackage{newunicodechar}
\usepackage{newunicodechar}
\newunicodechar{≈}{\approx}
\usepackage{booktabs}
\usepackage{graphicx}
\usepackage{wrapfig} 
\usepackage{siunitx}
\usepackage[utf8]{inputenc}
\usepackage{textgreek}
\usepackage[acronym]{glossaries}  
\usepackage{comment}

\setacronymstyle{long-short}
\newacronym{dvs}{DVS}{Dynamic Vision Sensors}
\newacronym{dbscan}{DBSCAN}{Density-Based Spatial Clustering of Applications with Noise}
\newacronym{rmse}{RMSE}{Root-Mean-Squared Errors}
\newacronym{ftir}{FTIR}{Frustrated Total Internal Reflection}
\newacronym{dnn}{DNN}{Deep Neural Networks}
\newacronym{rl}{RL}{Reinforcement Learning}
\newacronym{msows}{MSOWS}{Multitouch Soft Optical Waveguide Skin}
\newacronym{pdms}{PDMS}{Polydimethylsiloxane}
\newacronym{nir}{NIR}{Near-infrared}
\newacronym{sota}{SOTA}{State of the Art}
\newacronym{abs}{ABS}{Acrylonitrile Butadiene Styrene}
\newacronym{snn}{SNN}{Spiking Neural Network}
\newacronym{cmre}{CMRE}{Contact Map Reconstruction Error}
\newacronym{cusum}{CUSUM}{Cumulative Sum}
\newacronym{roc}{ROC}{Receiver Operating Characteristic}
\newacronym{tpr}{TPR}{True Positive Rate}
\newacronym{dl}{DL}{Deep Learning}
\newacronym{dcnn}{DCNN}{Deep Convolutional Neural Network}
\newacronym{eit}{EIT}{Electrical Impedance Tomography}
\newacronym{cnn}{CNN}{Convolutional Neural Network}
\newacronym{pet}{PET}{Polyethylene Terephthalate}
\newacronym{mems}{MEMS}{Micro-Electro-Mechanical Systems}

\def\keyFont{\fontsize{8}{11}\helveticabold }
\def\firstAuthorLast{Koolani {et~al.}}

\def\Authors{%
  Mohammadreza Koolani\,$^{1,*}$,
  Simeon Bamford\,$^{1}$,
  Petr Trunin\,$^{2,3}$,
  Simon F. M{\"u}ller-Cleve\,$^{1}$,
  Matteo Lo Preti\,$^{5}$,
  Fulvio Mastrogiovanni\,$^{4}$,
  Lucia Beccai\,$^{2}$, and
  Chiara Bartolozzi\,$^{1}$
}

\def\Address{%
  $^{1}$Event‐Driven Perception for Robotics, Istituto Italiano di Tecnologia, Genoa, Italy\\
  $^{2}$Soft BioRobotics and Perception Lab, Istituto Italiano di Tecnologia, Genoa, Italy\\
  $^{3}$The Open University Affiliated Research Centre at Istituto Italiano di Tecnologia (ARC@IIT), Istituto Italiano di Tecnologia, Genova, Italy \\
  $^{4}$ TheEngineRoom, Department of Informatics, Bioengineering, Robotics and Systems Engineering, University of Genoa, Italy \\
  $^{5}$ Advanced Robotics Centre, Department of Mechanical Engineering, National University of Singapore, Singapore 117575, Singapore
  }

\def\corrAuthor{Mohammadreza Koolani}
\def\corrEmail{Koolani.mohammad@gmail.com}

\begin{document}
\onecolumn
\firstpage{1}

\title[Neuromorphic Tactile Skin]{An Event-Based Opto-Tactile Skin} 

\author[\firstAuthorLast ]{\Authors} 
\address{} 
\correspondance{} 

\extraAuth{}

\maketitle

\begin{abstract}

This paper presents a neuromorphic, event-driven tactile sensing system for soft, large-area skin, based on the Dynamic Vision Sensors (DVS) integrated with a flexible silicone optical waveguide skin.
Instead of repetitively scanning embedded photoreceivers, this design uses a stereo vision setup comprising two DVS cameras looking sideways through the skin.
Such a design produces events as changes in brightness are detected, and estimates press positions on the 2D skin surface through triangulation, utilizing Density-Based Spatial Clustering of Applications with Noise (DBSCAN) to find the center of mass of contact events resulting from pressing actions. 
The system is evaluated over a 4620~\unit{\square\milli\metre} probed area of the skin using a meander raster scan. 
Across 95~\unit{\percent} of the presses visible to both cameras, the press localization achieved a Root-Mean-Squared Error (RMSE) of 4.66~\unit{\milli\meter}.
The results highlight the potential of this approach for wide-area flexible and responsive tactile sensors in soft robotics and interactive environments. 
Moreover, we examined how the system performs when the amount of event data is strongly reduced. 
Using stochastic down-sampling, the event stream was reduced to 1/1024 of its original size. 
Under this extreme reduction, the average localization error increased only slightly (from 4.66~\unit{\milli\metre} to 9.33~\unit{\milli\metre}), and the system still produced valid press localizations for 85~\unit{\percent} of the trials. 
This reduction in pass rate is expected, as some presses no longer produce enough events to form a reliable cluster for triangulation. 
These results show that the sensing approach remains functional even with very sparse event data, which is promising for reducing power consumption and computational load in future implementations. 
The system exhibits a detection latency distribution with a characteristic width of 31~\unit{\milli\second}.

\tiny
\keyFont{
\section{Keywords:} Tactile Sensing, Dynamic Vision Sensors, Optical Skin, Soft Robotics, Neuromorphic Engineering, Event-Based Sensing, Stereo Vision}
\end{abstract}

\section*{Data and Code Availability}

The dataset used in this study is available at Zenodo:~\url{https://zenodo.org/records/17274644}. 

The code for processing and analysis is available on GitHub:~\url{https://github.com/event-driven-robotics/optoskin}.

\section{Introduction}
Soft tactile sensors detect and interpret mechanical stimuli by leveraging compliant materials' inherent flexibility and adaptability. 
In the past few years, various transduction mechanisms have been extensively explored, including piezoresistive, capacitive, inductive, and optoelectronic approaches~\citep{Bartolozzi2016}. 
Among these, optical sensing stands out due to its wide sensitivity range, excellent reliability, and inherent resistance to electromagnetic interference~\citep{LoPreti2022chapter}. 
Unlike many tactile sensors relying on dense wiring or embedded electronics within the sensing area, non-array soft optical waveguides benefit from a sensitive area devoid of wires and rigid parts, delivering fully flexible and adaptable sensing surfaces~\citep{Kamiyama2004, Li2022}. 
This feature mainly benefits tactile systems requiring conformity to complex or changing surfaces~\citep{Chorley2009, Guo2022b}. 
Moreover, spatial data in optical soft sensors can be retrieved from broad regions without embedded electronics, relying on data processing, which supports seamless integration into robot platforms and versatile applications in environments that are dynamic, unstructured, or subject to frequent change.~\citep{Guo2022b, Faris2023}. 

Two main types of optical tactile sensors can be identified, namely Frustrated Total Internal Reflection (FTIR) and vision-based systems. 
On the one hand, FTIR sensors operate by detecting changes in light propagation within an optical waveguide caused by deformation or contact, thus allowing a precise measurement of pressure and shape through photodetection~\citep{Trunin2025, LoPreti2025}. 
Vision-based optical sensors, on the other hand, utilize cameras to capture deformation or changes in the sensor surface~\citep{Shimonomura2019}. 
Vision-based optical sensors, which capture deformation or surface changes via cameras, can be divided into frame-based and event-based approaches.
Frame-based vision tactile sensors typically capture images at fixed rates to monitor surface deformation or marker displacements, enabling spatially detailed force and shape sensing. 
For example, \cite{Kamiyama2004} use color-coded markers embedded in a transparent elastic layer to map three-dimensional contact forces, aiding delicate robot tasks. 
The compact DIGIT sensor~\citep{Lambeta2020} implements low-cost, high-resolution tactile feedback with a modular elastomer design suitable for robust in-hand manipulation and industrial use. 
Alternative optical methods, such as the retrographic sensor by~\cite{Johnson2009} leverage photometric stereo to detect surface texture and shape, while~\cite{Trueeb2020} employ multiple cameras and adopts \gls{dl} for 3D force reconstruction over larger areas, enabling scalable robot skins. 
Inspired by human fingertip anatomy, the TacTip sensor~\citep{Chorley2009} uses internal marker displacement within an artificial papillae structure to achieve sensitive edge detection and manipulation without any embedded electronics.

\gls{dvs}---also known as event cameras---are inspired by biological vision systems and detect brightness changes asynchronously, providing high-speed, low-latency, and data-efficient sensing~\citep{Lichtsteiner2008, Gallego2019}. Event-based approaches have driven advances in tactile perception for robots.
Event-based approaches have driven advances in data-efficient tactile perception for robots. Recent reviews highlight that neuromorphic tactile sensors emulate biological mechanoreceptors using spike-based, event-driven encoding, producing sparse, low-latency, and energy-efficient touch signals~\cite{Liu2025, Kang2024}.
For example, \cite{Faris2023} integrate \gls{dvs} cameras into a soft robotic finger, achieving proprioception and slip detection via real-time event-based heat maps with a 2~\unit{\milli\second} latency, enhanced by \gls{dl} techniques and a flexible fin-ray finger design.
Similarly, \cite{Naeini2020} demonstrates a successful application of \gls{dvs} sensors for high-sensitivity force estimation and material classification, highlighting reduced computational and energy demands compared to frame-based methods in dynamic scenarios. 
The hybrid DAVIS sensor~\citep{Brandli2014}, which generates as output both asynchronous events and synchronous frames, is employed by~\cite{Rigi2018} with silicone substrates to detect slip and vibration, therefore showcasing suitability for rapid, high-resolution tactile feedback. 
Building on marker-based tactile sensing, NeuroTac combines the TacTip design~\citep{Lepora2021} with \gls{dvs} sensors to enable object recognition, shear detection, grasp stabilization, and manipulation for anthropomorphic robots~\cite{WardCherrier2020}. 
In a complementary approach, \cite{Sferrazza2019} developed a high-resolution optical sensor using marker tracking and \gls{dl} for real-time force prediction on soft surfaces. 
More recently, \cite{Funk2024} introduced Evetac, an event‑based optical tactile sensor leveraging 1000~\unit{\hertz} event readout rate to sense vibrations up to 48~\unit{\hertz}, reconstruct shear forces, drastically reduce data rates versus RGB sensors, and demonstrate data‑driven slip detection within a robust closed‑loop grasp controller. In a complementary approach, \cite{Sankar2025} demonstrate a biomimetic prosthetic hand with three layers of neuromorphic tactile sensors, enabling compliant grasping and achieving $\approx$99.7\unit{\percent} accuracy in texture discrimination.

The approaches mentioned above focus on small-scale sensors for fingertips, with limited examples of large-area tactile skins. 
Additionally, many of these solutions rely on high-resolution vision sensors, which often require substantial computational resources and the use of advanced data-driven techniques, such as \glspl{dnn} or \gls{rl}, to process the vast amount of generated data. 
While promising, these approaches may face scalability, energy efficiency, and real-time processing challenges, especially for large-scale, adaptive soft robot systems. As an extreme example of multimodal touch sensing, \cite{Meta2024} built an artificial fingertip with approximately 8.3 million taxels capable of capturing fine pressures, high-frequency vibrations (up to 10~\unit{\kilo\hertz}), thermal cues, and chemical traces, all processed by on-chip AI.

To the best of our knowledge, we present the first system integrating stereo \gls{dvs} sensors with an optical skin to perform contact localization via triangulation. 
Unlike prior works mostly emphasizing force estimation on small fingertip areas, our approach provides a lightweight solution for the localization of contacts, scalable to large flexible skin areas, using minimal computation without relying on either \glspl{dnn} or complex marker tracking. 
However, contact‐point localization has also been studied acoustically, for example, \cite{Lee2024} introduced ``SonicBoom,'' in which a distributed microphone array mounted on a robot link pinpoints single taps with sub‑centimeter accuracy even under occlusions.

This study builds on~\cite{LoPreti2022}'s \gls{msows}, a marker-free, optical skin that achieves large-area tactile sensing without embedded electronics in the sensing area. 
The design combines graded-stiffness \gls{pdms} layers and a \gls{nir} optical system to sense pressure, providing adaptability to complex shapes and durability against repeated mechanical stress. 
\gls{msows} uses pulsed light sources and repetitive sequential scanning through photodiodes integrated around the skin perimeter.
The rate of this scan introduces inherent latency into the detection process, which can only be reduced by increasing the scan rate and therefore the volume of generated data.
In this work, we leverage the low-latency change detection capabilities of event-driven photoreceptors to detect the changes in illumination due to tactile contacts on the silicone surface. 
As a first step to substitute the array of photodiodes in~\cite{LoPreti2022}, we conducted a preliminary study using a pair of \gls{dvs} integrated in the silicone layer for contact localization.  

As a potential skin covering, applicable to larger areas, this setup allows for an efficient triangulation and spatial analysis of touch positions, therefore contributing to the development of scalable and adaptive tactile sensors in soft robots. 
Unlike the above-mentioned approaches, which rely on \glspl{dnn} or heavy data-driven algorithms, this work uses a simple, lightweight triangulation for localizing pressure locations on the silicone surface, which is crucial for real-time applications.

An overview of the performance metrics of \gls{sota} optical tactile sensors is summarized in Table~\ref {tab:comparison}, comparing with our approach and showing their hardware, the measurements, the algorithms, and the task-relative metrics.

\begin{table*}[!t]
  \centering
  \caption{Comparison of this work and similar tactile sensors}
  \label{tab:comparison}
  \begin{tabular}{p{3cm} p{3cm} p{3cm} p{3cm} p{4cm}}
      \toprule
      \textbf{Sensor} & \textbf{Hardware} & \textbf{Primitive} & \textbf{Algorithm} & \textbf{Metrics} \\
      \midrule
       Force-EvT (\cite{Guo2024}) & 
      DVS & 
      Force estimation & 
      Vision Transform & 
      RMSE 0.13~\unit{\newton} (Newtons) \newline R$^2$=0.3 \newline 1.5~\unit{\percent} error\\
      
      \addlinespace
      
      Vision-Based (\cite{Naeini2020}) & 
      DVS & 
      Contact force & 
      Time Delay Neural Network & 
      RMSE 0.16--0.17~\unit{\newton} \newline 7.2~\unit{\percent} accuracy \\
      
      \addlinespace
      
      Bio-Inspired (\cite{Faris2024}) & 
      DVS & 
      Slip \& Pressure & 
      Spike processing & 
      5~\unit{\percent} accuracy \newline 2~\unit{\milli\second} RT \\

      \addlinespace
      
      Evetac (\cite{Funk2024}) & 
      DVS & 
      Vibration \& Shear force & 
      \gls{dl} & 
      Vibration sensing \newline up to 48~\unit{\hertz} \\

      \addlinespace
      
      E-BTS (\cite{Mukashev2025}) & 
      DVS & 
      Force estimation & 
      \gls{dl} force prediction & 
      RMSE $\sim$1~\unit{\newton} \newline Latency $<$ 5~\unit{\milli\second} \\

      \addlinespace
      
        NeuroTac (\cite{WardCherrier2020}) &
        DVS &
        Texture / Object recognition, slip cues &
        Spiking neural network (event-based) &
        Classification accuracy (task-dependent; e.g. ~95\unit{\percent} on Braille) \\

      \addlinespace
      
      Multi-Cam (\cite{Trueeb2020})& 
      4 RGB cameras with silicone & 
      Force distribution & 
      \gls{dcnn} & 
      RMSE 0.057~\unit{\newton} (Fz) \newline 40~\unit{\hertz} \\

      \addlinespace
      
      DIGIT (\cite{Lambeta2020})& 
      RGB camera with elastomer & 
      Shape \& force & 
      Autoencoder & 
      640×480 px \newline @ 60~\unit{\hertz} \\

      \addlinespace
      
      Retrographic (\cite{Johnson2009}) & 
      RGB with elastomer & 
      Texture/Shape & 
      Photometric stereo & 
      High-res 2.5D capture \\

      \addlinespace
      
      MSOWS (\cite{LoPreti2022}) & 
      LED with photoreceiver & 
      Localization & 
      Sequential scan & 
      5~\unit{\milli\meter} resolution \newline 2.6~\unit{\percent} error \\

      \addlinespace
      
      This Work & 
      Stereo DVS & 
      Contact location & 
      Stereo with DBSCAN & 
      RMSE 4.66~\unit{\milli\meter} \newline 31~\unit{\milli\second} detection latency distribution \\
      
      \bottomrule
  \end{tabular}
\end{table*}

\section{Materials and Methods}

\subsection{Sensor Design and Manufacturing Process}\label{design}

This paper introduces a new design for a vision-based sensor using two \gls{dvs} cameras, partly inspired by classic soft optical sensors integrating LEDs and receivers within soft, transparent silicone substrates. 
This sensor features a square 100~\unit{\milli\meter} per side and 4~\unit{\milli\meter} thick silicone layer, embedding two \gls{dvs} cameras (Prophesee EVK1 VGA Gen3.1 sensor, resolution: $u = 640$, $v = 480$), positioned at the corners of one edge. 
Furthermore, 32 \gls{nir} LEDs are mounted along the remaining three edges, all pointing to the center of the edge with the two cameras. 
This setup helps enhance the visibility of the pressure distribution applied to the silicone layer.
The LEDs emit light into the elastic silicone element, where deformation within the material affects the light intensity, which is then detected by the \gls{dvs} cameras. 
Using two cameras allows for stereopsis, therefore enabling a 2D localization of pressure locations on the skin.

\begin{figure}[H]
    \centering
    \includegraphics[width=0.6\textwidth]{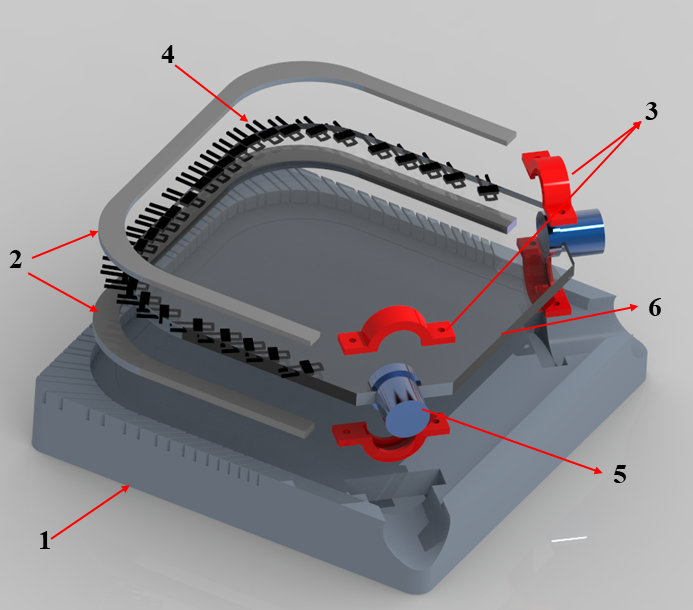}
    \caption{Exploded view of the sensor design. Components: (1) ABS printed base; (2) emitter mounts; (3) lens mounts; (4) near-infrared VSMY1860 emitters; (5) ABS lens replicas used during molding; (6) PDMS tactile layer.}
    \label{cad_model_of_the_sensor}
\end{figure}

The sensor design is shown in Fig.~\ref{cad_model_of_the_sensor}. 
The base~(1), printed from \gls{abs} on an Ultimaker S3 (Ultimaker, Utrecht, NE), houses mounts for the emitters~(2) and camera lenses~(3). 
\gls{nir} VSMY1860 emitters~(4) and \gls{abs}-printed lens replicas~(5) are placed into these mounts. 
3D-printed dummy lenses were used in the silicone casting process, allowing for easy removal without damaging the silicone layer. 
Those replicas are then replaced with actual lenses afterward.

After assembling all components, silicone~(6) was prepared and deposited in the mold. 
Due to its well-known optical properties, particularly its optical transmittance, \gls{pdms} was chosen as the elastic element. 
A 30:1 ratio of \gls{pdms} base to curing agent was prepared and mixed using a Thinky Mixer (THINKY U.S.A., INC.), then degassed in a vacuum chamber for 7~\unit{\minute} before being poured into the mold. 
This ratio was chosen to make the \gls{pdms} as soft as possible, enhancing the sensor's sensitivity and flexibility~\cite{Sales2021}. 
This softer composition facilitates the detection of subtle interactions and deformations, improving its responsiveness and the overall performance. 
Finally, the sensor was cured in an oven for 2~\unit{\hour} at 75~\unit{\celsius}.

\subsection{Experimental Setup and Data Acquisition}
\label{sec:exp_set}

\begin{figure}[H]
    \centerline{\includegraphics[width=0.4\textwidth]{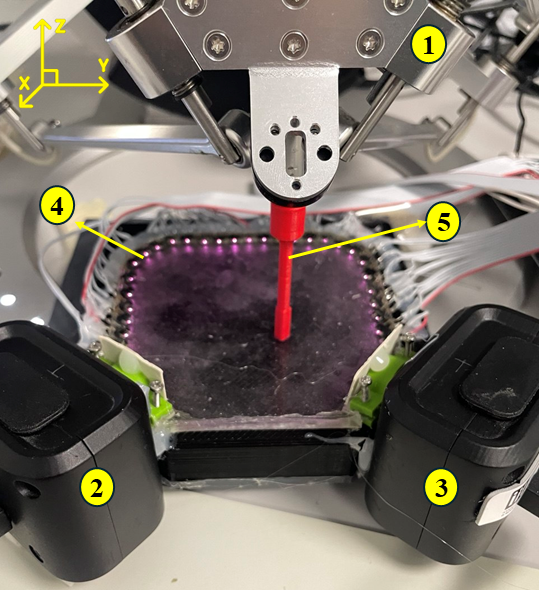}}
    \caption{Experimental setup for tactile data acquisition with our optical skin system. The setup includes DVS cameras~(2 and 3) placed on two adjacent sides of a rectangular silicone layer to capture dynamic touch events. The Omega~3 force dimension robot~(1) applies a controlled force using the tip~(5) to predefined points on the silicone surface, creating deformations detected by the cameras. NIR LEDs~(4) provide consistent illumination, allowing detection of surface deformations.}
    \label{exp_set}
\end{figure}

Data acquisition in the setup shown in Fig.~\ref{exp_set} is carried out by applying an external force to the silicone layer using an Omega~3 Force Dimension robot. 
This leads to deformation and brightness changes detectable by the \gls{dvs} cameras. 
The sensing area is rectangular, with \gls{dvs} cameras at two corners. 
This arrangement covers most of the tactile sensing area, allowing each camera to capture variations in brightness caused by the deformation of the silicone layer illuminated by \gls{nir} LEDs placed on the other sides, as illustrated in Fig.~\ref{cad_model_of_the_sensor}. 
The \gls{nir} LEDs (16~\unit{\milli\watt} total power consumption) provide consistent illumination, thus ensuring high visibility of the silicone layer even in low-light conditions, enhancing the accuracy of touch-point detection. 
This configuration is expected to support efficient and highly dynamic range data acquisition, enabling responsive and adaptable sensing in robots and haptic settings.

The sensor was pressed at 250 locations arranged in a meander path with 4~\unit{\milli\meter} spacing between each point at a constant depth of 2~\unit{\milli\meter} (it should be mentioned that the applied force was not analyzed). 
To ensure data consistency and robustness, the grid pressing pattern was repeated 10~times, each involving the same 250 pressing points. 
After completing each set of 250 presses, the robot returns to the starting position to begin the next repetition.

To ensure temporal alignment between the two \gls{dvs} recordings, a synchronization procedure is performed prior to the main experiment. 
Since the recordings were started manually and were not inherently synchronized, the center of the skin---clearly visible to both cameras---was pressed three times with 1-\unit{\second} intervals, followed by a 3-\unit{\second} pause before the first actual press. 
These three initial press actions serve as temporal landmarks, enabling a precise alignment of timestamps across both camera streams. 
The aligned event sequences ensure the accurate segmentation and a consistent comparison of press timings. 
These initial calibration actions are highlighted in Fig.~\ref{eventrates}~A with a rectangle.

\begin{figure}[htbp]
    \centering
    \includegraphics[width=\textwidth]{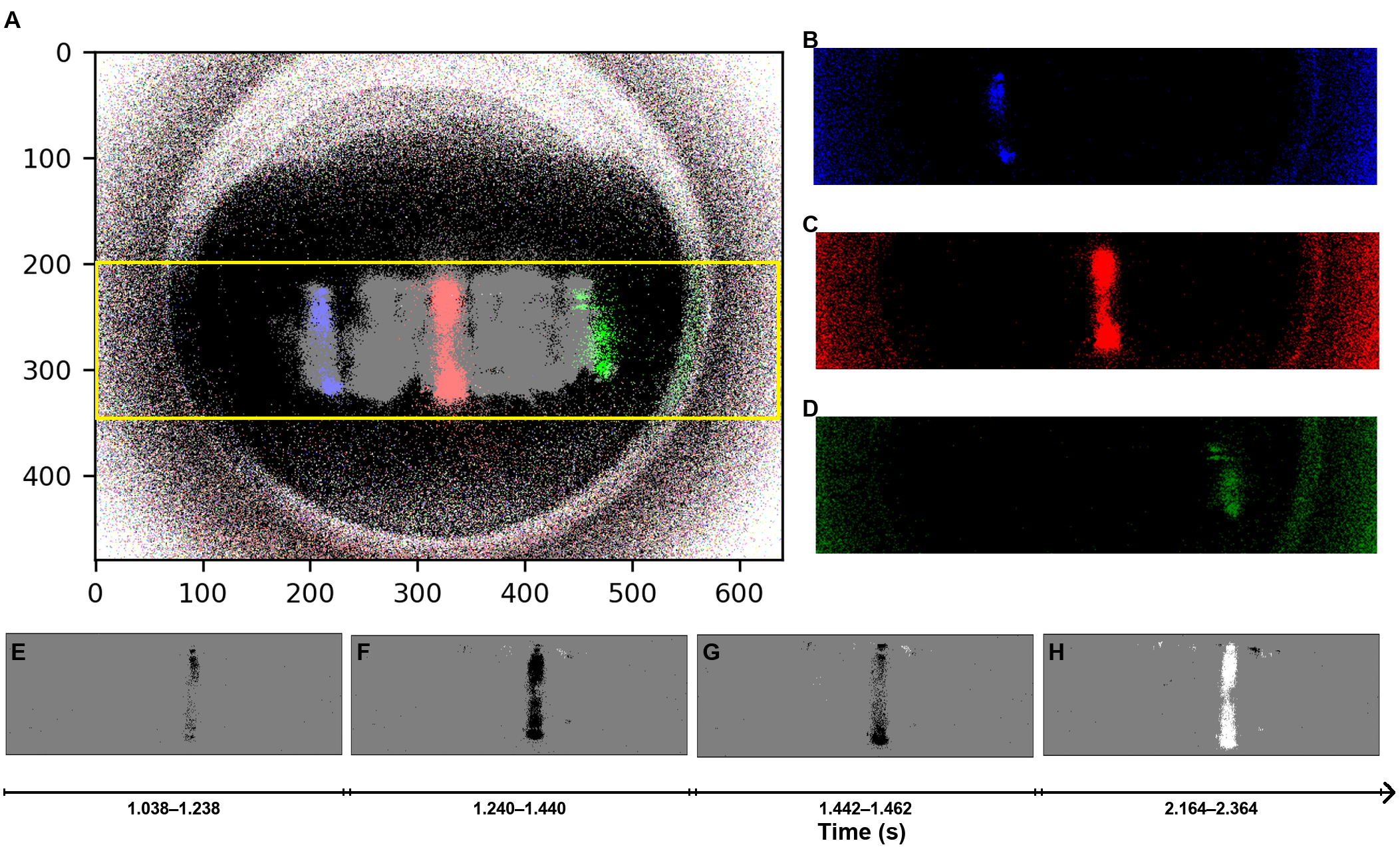} 
    \caption{Overview of DVS camera event data illustrating press actions localization based on the circular path. Image~A shows the accumulated event rates across all contacts, providing a high-level view of spatial activity distribution over the sensing area. The yellow box indicates the cropped region (v = 200--360), and only the pixels inside this box are retained for analysis. Images~B, C, and D represent individual contacts captured by the same camera, each shown in a color-coded channel. These images demonstrate the sensor's sensitivity to different press locations, as each press generates a unique pattern of event distribution. Images~E, F, G, and H (right column) display a temporal sequence for a single contact, highlighting spatiotemporal characteristics that could be useful for further classification.}
    \label{cal_before}
\end{figure}

The accumulated event rates for contacts on the circular path (48 presses on two circles with radii of 20~\unit{\milli\metre} and 25~\unit{\milli\metre}) are shown in Fig.~\ref{cal_before}~A. This dataset is used solely to illustrate how press-induced event activity appears on the sensor surface, as the circular pattern provides visually clear and evenly distributed examples of event generation. No localization analysis is performed on this dataset. All quantitative results presented in this work---including triangulation accuracy, pass-rate analysis, and detection latency distribution evaluation---are based exclusively on the meandering-path dataset shown in Fig.~\ref{eventrates} and~\ref{centroids}.

Much of the sensing area remains unused, as seen by the outer ring around the viewable area. 
The images in Fig.~\ref{cal_before}~B--D represent individual contacts on the same camera, each shown in a different color. 
These images highlight the varying activity levels across different pressure locations, illustrating how each press action generates a distinct pattern of event distribution on the sensor.

Each contact captured by the \gls{dvs} cameras exhibits unique spatio-temporal characteristics, as illustrated in the sequence of images in Fig.~\ref{cal_before}~E--H on the right, which show the stream of events generated from a single press with events accumulated over a fixed time window. 
In these images, gray pixels represent areas without events, while black and white pixels indicate events with negative and positive polarities, respectively, corresponding to decreases and increases in light intensity detected by the sensor. 
This reflects how light from LEDs is blocked or revealed during skin deformation as the indenter presses and releases. 
As illustrated in Fig~\ref{exp_set}, the cameras are embedded at the edge of the skin, looking through the side of the silicone. 
The events are generated by the indenter pressing the top surface of the skin, causing deformations and interrupting the light coming from the LEDs visible from both cameras. 
This temporal progression of events highlights changes in the activity and could be leveraged for further processing, extracting additional information from the data, such as the classification of contact types, slip detection, or motion patterns. 
However, this study is focused on the localization of contacts based on aggregated event statistics.

\subsection{Data Processing and Press Actions Segmentation}
\begin{figure*}[htbp]
    \centerline{\includegraphics[width=\textwidth]{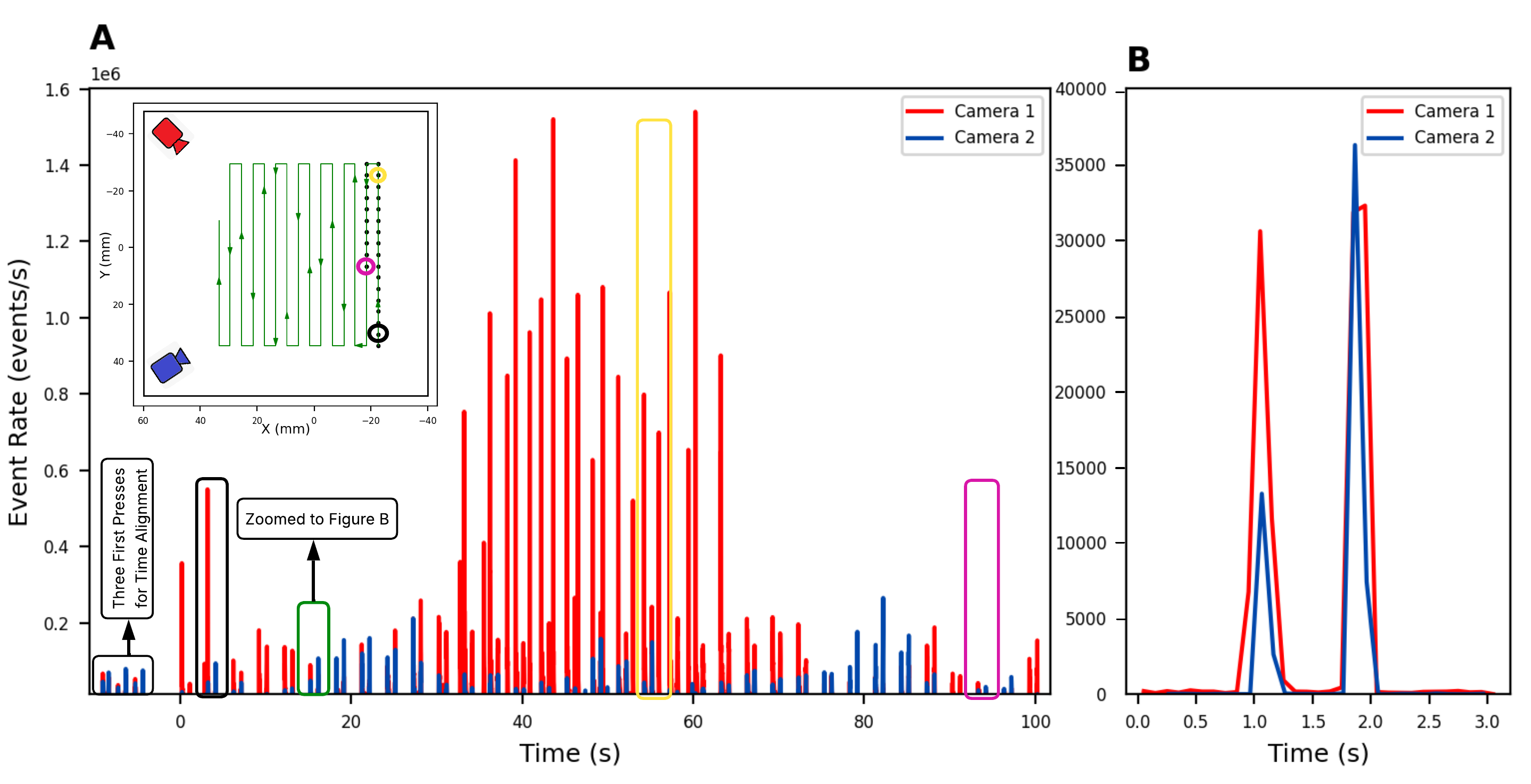}}
    \caption{Event rate analysis during skin pressing, computed using histograms with a bin size of 10~\unit{\milli\second}. 
    (A) Event rate over the first 100~\unit{\second} of activity from Camera~1 (red) and Camera~2 (blue), corresponding to the first 27 presses performed in a meandering path. The inset shows the sensor area and pressing path, with camera positions indicated (red and blue) and individual presses marked. Selected presses are highlighted with color-coded rectangles on the event-rate plot and linked to circles of the same color on the sensor map: green for the 2nd press, yellow for the 16th press, and purple for the 27th press. 
    (B) Event-rate for a single pressure instance recorded by both cameras, overlaid to show synchronization and comparative sensitivity. Camera~1 is plotted in red and Camera~2 in blue consistently.}
    \label{eventrates}
\end{figure*}

The \gls{dvs} cameras generate continuous event data, with each press action on the optical skin generating a series of events. 
Initial data processing begins by calculating the event rate (number of events per second) across all press actions, offering a high-level overview of tactile activity over time.  Fig.~\ref{eventrates}~A displays the event rate from both Camera~1 (red) and Camera~2 (blue) over the first 100~\unit{\second} of the experiment. 
This corresponds to 27 individual pressures applied in a meandering path, as illustrated in the inset on the top left. 
The inset also shows the spatial layout of the sensor and the robot's pressing path, with the relative positions of the two cameras (red and blue) and the pressure trajectory over the optical skin.
Each contact location is marked, and selected pressures are connected to their corresponding spikes in the event-rate plot using green rectangles for visual reference.

To remove irrelevant activity and focus only on tactile interactions, the upper and lower part of the visual field was cropped (on the v-axis, from $v = 200$ to $v = 360$), based on the 4~\unit{\milli\meter} thickness of the silicone layer and the lens characteristics. 
The cropped region corresponds to the yellow box shown in Fig.~\ref{cal_before}~A.

Furthermore, to enable press-wise analysis, event data were segmented temporally into individual pressure events. 
Fig.~\ref{eventrates}~B zooms in on a single contact and shows the corresponding event-rate for Camera~1 (red) and Camera~2 (blue), highlighting synchronization and sensitivity of both views during a single tactile event.

\subsection{Event Detection and Finding the Centroids}
After segmenting the individual pressures, we used the clustering-based approach \gls{dbscan} to find the horizontal centroid of each contact on the sensor of each camera.

\gls{dbscan} identifies the density of the clusters of contact activity without prior assumptions about the number of events or the shape of the clusters. 
As all presses are separated into individual containers based on their timing, \gls{dbscan} can be applied to spatial data to find the region with maximum activity. 
Events that do not belong to any cluster are excluded as noise. DBSCAN was applied with an $\varepsilon$ (eps) neighborhood of 10 pixels and a minimum of 10 events per cluster ($min\_samples = 10$). These parameters were chosen empirically to balance noise rejection with robust cluster formation across different presses.

When one or more clusters are found, the largest cluster is computed and considered as the direction of the contact activity on the sensor’s surface. 
This method filters out scattered noise events and focuses on the activity elicited by pressure contacts. 
From the remaining events, the centroid of the dominant cluster is computed as the mean of the $(u, v)$ positions within the cluster. 
This centroid is used to represent the pressure location in the image space.
Although both $u$- and $v$-coordinates are used for clustering, only the $u$-coordinates of the resulting centroids are retained for subsequent triangulation Sec~\ref{sec:triangulation}. 
This is because the two cameras are arranged along a horizontal stereo baseline, meaning that disparity, and consequently depth information, is primarily encoded in the horizontal ($u$) direction. 
The $v$-coordinate variations do not contribute to the triangulation accuracy in this configuration but remain useful for verifying the spatial consistency of the detected contact regions.
The process is repeated for all presses across both cameras.

\begin{figure*}[htbp]
    \centerline{\includegraphics[width=\textwidth]{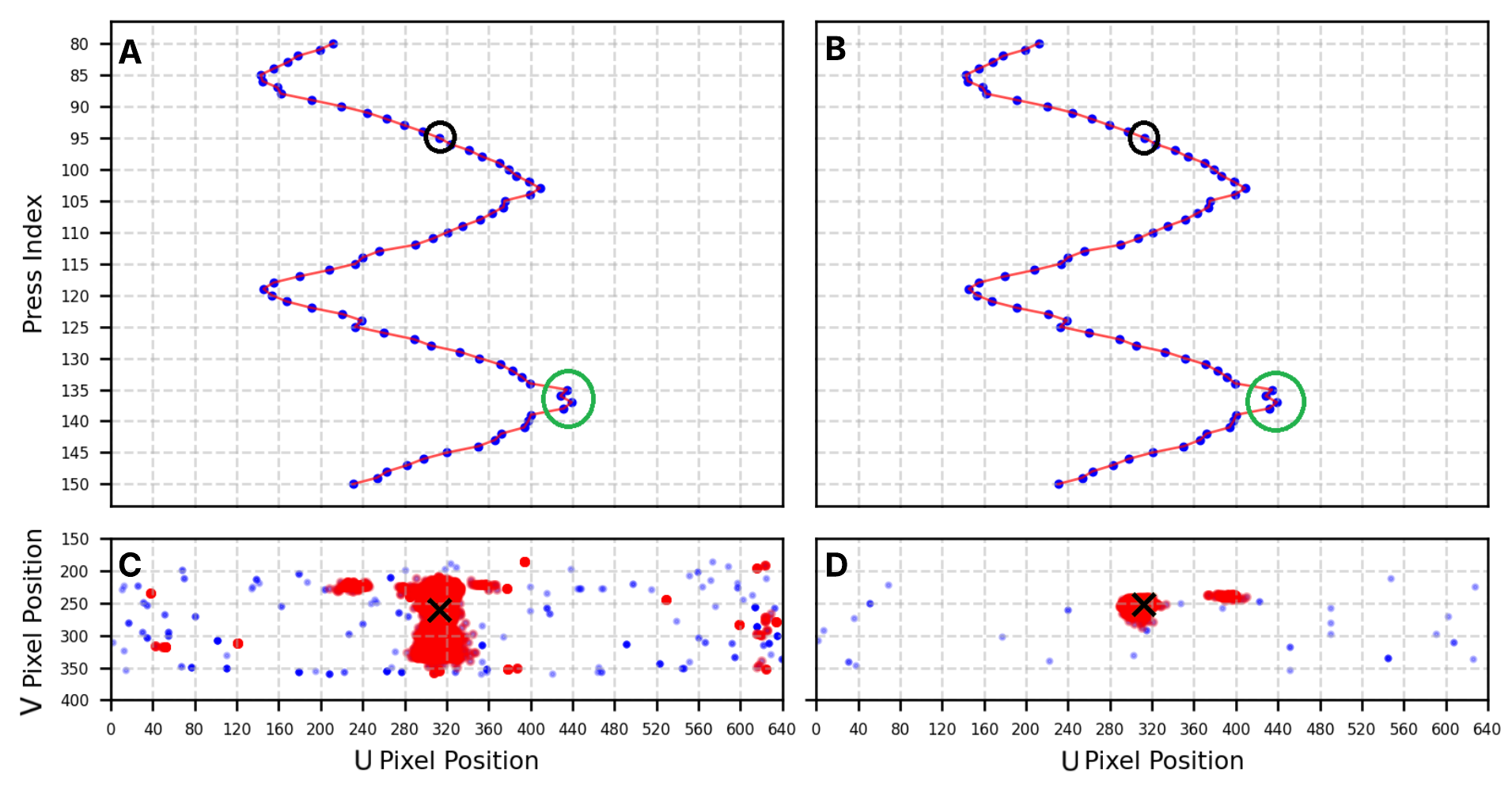}}
    \caption{The spatial distribution of pixel activity using DBSCAN. This figure visualizes the estimated activity centers across the sensor surface based on DBSCAN clustering of pixel events. (A) and (B) show the top-down views of the $u-$coordinate centroids extracted from each press for Camera~1 and Camera~2, respectively. These views illustrate the spatial spread of detected presses along the circular path. (C) and (D) show example contact activity for Camera~1 and Camera~2, where the DBSCAN-identified pixel clusters have been processed to extract their centroids, representing the dominant activity locations. In these examples, the clusters correspond to press number~95, which is marked with a black circle in subplots~A and~B, and the black cross in subplots~C and~D indicates the center of the activity for that press.}
    \label{centroids}
\end{figure*}

Fig.~\ref{centroids}~A~and~B present the top-down distribution of centroid $u-$positions identified by \gls{dbscan} for contacts numbered from 80 to 150, visualized from the perspectives of Camera~1 and Camera~2, respectively. 
This visualization offers a mapping of detected pressure locations across the sensor surface.
Additionally, Fig.~\ref{centroids}~C~and~D show individual contact activity distributions with their clustered activity regions.

\subsection{Data Filtering and Pressures Exclusion Criteria}
Fig.~\ref{centroids} illustrates scenarios in which press localization may fail due to weak or ambiguous event patterns in either view.
Such weak pressure activities are highlighted by green circles in subplots~A and~B. 
To ensure a robust localization, a press was removed if \gls{dbscan} failed to identify a prominent cluster of activity in one of the views. 
This happens when there is not enough density or structure in the spatial arrangement of events to create a significant cluster, which might indicate that the pressure was too weak or was obscured.
A threshold number of cluster points was introduced as a metric to determine the existence of a valid pressure event. 
To perform the optimization in Sec.~\ref{sec:triangulation}, some outliers were manually excluded (this is consistent with a one-off calibration process for a new sensor). 
However, the outliers remaining after thresholding were included. 

\subsection{Geometric Triangulation for Pressure Localization} 
\label{sec:triangulation}

Once the area of maximum activity for the $u-$coordinate of each camera is found, the angle between the center pixel of each camera’s field of view and the identified active pixel (notated as $\theta_1$ for Camera~1 and $\theta_2$ for Camera~2) is computed. 
These angles, along with the known distances (that is, $d_1$ and $d_2$) from each camera to the center of the silicone layer, allow us to draw triangulation lines from both cameras toward the detected pressure location. 
Angles and distances are inputs for geometric triangulation, providing the basis to estimate the press location on the sensor’s $x-$ and $y-$coordinates.

\begin{figure}[H]
    \centering
    \includegraphics[width=0.6\columnwidth]{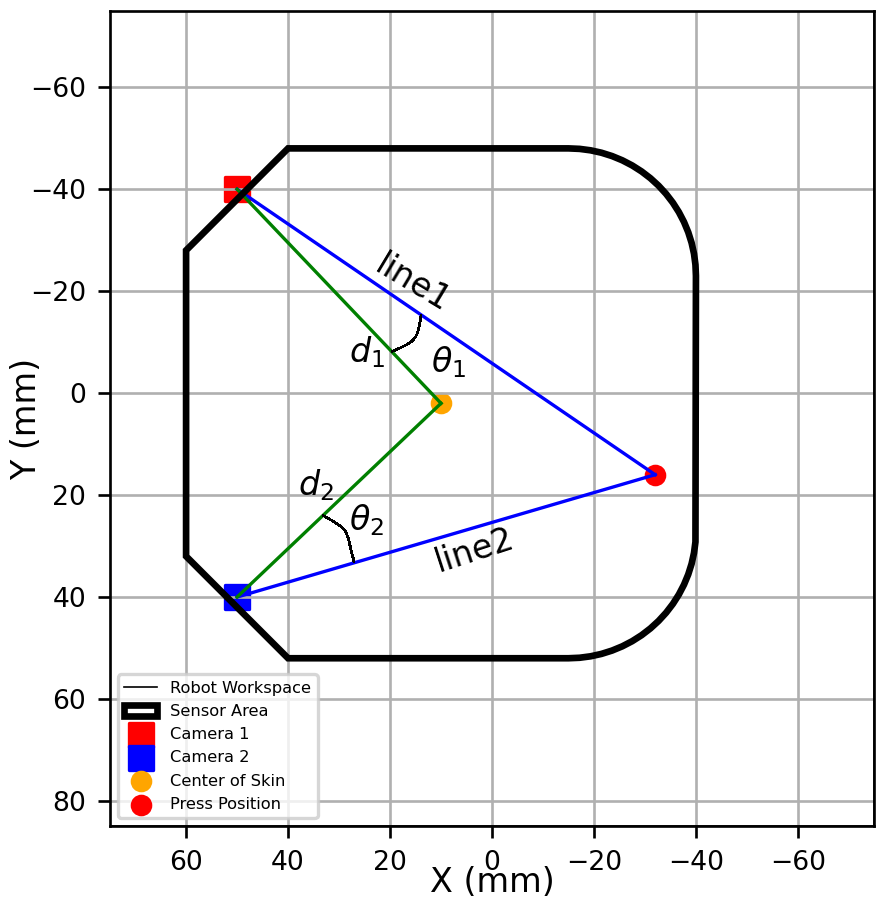} 
    \caption{The triangulation process for contact position estimation. This figure shows the triangulation process, where angles ($\theta_1$ and $\theta_2$) are calculated between each camera’s center and active pixels. Distances ($d_1$ and $d_2$) from the cameras to the center of the sensing area are used to draw lines towards the detected pressure position. The intersection of these lines provides the estimated coordinates of the press on the optical skin.}
    \label{triangulation}
\end{figure}

The final pressure position is determined by computing the intersection of the triangulation lines from each camera, representing the estimated coordinates $(x, y)$ of the press on the optical skin. \gls{dbscan} allows for the estimation of the activity center, effectively mitigating noise and identifying center regions of pixel activity. 
Fig.~\ref{triangulation} illustrates this triangulation process, detailing the angles and distances of locating each press in the sensing area.
Although initial triangulation was performed using camera positions measured directly from the robot workspace, further refinement was introduced to improve localization accuracy by treating the following system parameters as optimization variables.
Skew angles were introduced into the triangulation model to account for any possible misalignment between the cameras and the edges of the tactile surface. 
Recognizing that the cameras might not be perfectly parallel to the skin's edges due to setup imperfections, the skew angles and camera positions are treated as variables within the triangulation process. 
By incorporating skew angles ($\theta_{\text{skew}_1}$ for Camera~1 and $\theta_{\text{skew}_2}$ for Camera~2) and optimizing camera positions ($ x_1, y_1$ for Camera~1 and $x_2, y_2$ for Camera~2) and lens distortion, the accuracy of estimated press locations was improved. 
The least squares optimization was then performed on these parameters to minimize the difference between the estimated pressure locations and the coordinates provided by the Omega~3 force dimension robot. 
This optimization was performed for the grid press trajectories to account for any differences in the data patterns. 
Optimized camera positions, skew angles, and distortion correction improved accuracy, as reflected in the error metrics presented in Sec.~\ref{sec:error}. It should be noted that, as described in the experimental setup, the grid press path consisting of 250 locations was repeated ten times. The triangulation parameters were optimized using data from one repetition of the path, and the optimized parameters were then applied to the remaining nine repetitions. This approach was adopted to minimize overfitting and to provide a more reliable estimate of the system’s localization accuracy.

The differences in the optimized camera positions for the grid path are visible in Fig.~\ref{result}, where the camera positions are depicted differently for the path. 
Triangulation calculations are based on these introduced and optimized variables.

\section{Results}

This Section presents the performance of the proposed event-based optical skin in localizing contact events. 
We report the system’s localization accuracy based on ground truth comparisons, analyze its spatial coverage and detection robustness, and evaluate its response to data reduction through ablation testing. 
In addition, the potential latency distribution of tactile event detection is quantified to assess the sensor’s suitability for real-time applications.

\subsection{Error Calculation and Comparison} 
\label{sec:error}

\begin{figure}[H]
    \centering
    \includegraphics[width=0.6\columnwidth]{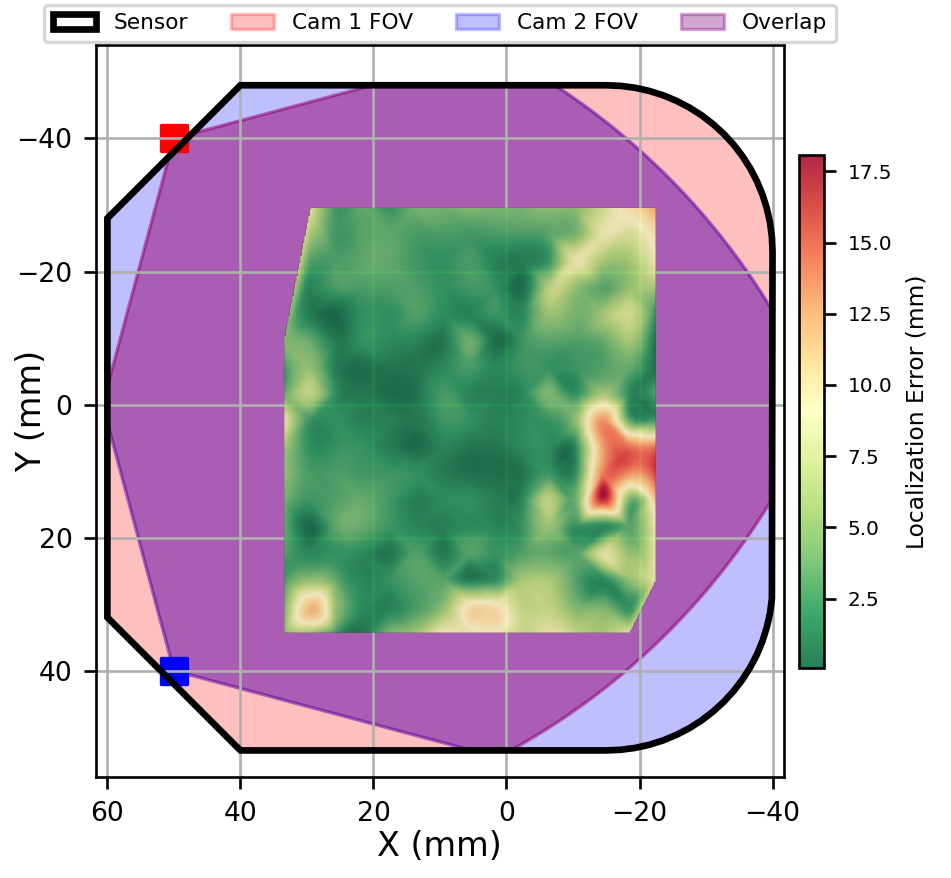} 
    \caption{Localization error heatmap for the DBSCAN-based triangulation method on the grid path. Red regions indicate higher error magnitudes, while green regions correspond to more accurate estimations. The field of view (FOV) of each camera, their overlap region, and the sensor boundary are also shown.}
    \label{result}
\end{figure}

To evaluate the accuracy of the triangulation method, the estimated pressure locations are compared to the ground-truth coordinates provided by the Omega~3 force dimension robot. 
The \gls{rmse} was selected as the primary performance indicator among the calculated error metrics.
The \gls{rmse} for the grid path is 4.66~\unit{\milli\metre}, which corresponds to 3.75~\unit{\percent} of the sensor’s diagonal. 
The mean standard deviation between trials for single pressures on the grid path is 2.85~\unit{\milli\metre}.
This indicates that the triangulation approach can estimate press positions, providing sufficient accuracy for practical tactile sensing applications. 
Fig.~\ref{result} illustrates the estimated pressure locations derived from \gls{dbscan} for the grid path.
To better understand how the localization error is distributed, we also calculated the RMSE separately along the $x$- and $y$-coordinates. The RMSE in the $x$-direction is 4.17~\unit{\milli\metre}, and the RMSE in the $y$-direction is 3.28~\unit{\milli\metre}. These values show how the triangulation geometry contributes differently along each axis and complement the overall Euclidean RMSE of 4.66~\unit{\milli\metre}.

The total area of the fabricated skin surface is 9328~\unit{\square\milli\metre}. 
Out of this area, 8763~\unit{\square\milli\metre} (93.9~\unit{\percent}) is within the field of view of both cameras. 
The robot probed 4620~\unit{\square\milli\metre} (52.7~\unit{\percent}) of the visible skin area. 
After applying a manual threshold to reject poor quality results, the proportion of probed locations, which are visible to both cameras, where pressures can be reliably detected and localized, is 95~\unit{\percent}.

\subsection{Data Ablation and Latency Distribution Analysis} 
One of the main motivations for this work is achieving a low-data sensor maintaining the accuracy of the localization. 
High-resolution \gls{dvs} cameras are utilized in the present realization. 
Subsequent system versions could use dedicated event sensors with substantially fewer pixels, consequently reducing data transmission and processing demands. 
Furthermore, optimization of sensor tuning would also enhance responsiveness, with better control over the event rate. 

To investigate how much the system depends on the amount of event data, we performed a data ablation (event reduction) analysis. Event data was progressively reduced using predefined downsampling factors ($2^0$ to $2^{10}$), and the localization accuracy was evaluated using the \gls{rmse} and pass rate metrics.

The downsampling was done by applying a simple stochastic thinning of the event stream. For a given reduction factor~$k$, each event was kept with probability $1/k$ and discarded otherwise. This keeps the original timing of the events but reduces their total number by roughly a factor of~$k$. For each reduction factor, we repeated the thinning process using different random seeds to check the consistency of the results.

For each press, we computed the Euclidean distance between the estimated and actual press coordinates. For a given reduction factor, the distribution of these errors was obtained over all presses. A press was counted as a ``pass'' if its localization error was smaller than the 95$^{th}$ percentile of that error distribution. The pass rate is the proportion of presses that meet this criterion. This gives an intuitive measure of how often the system still produces a usable localization after reducing the number of events.

\begin{figure}[H]
    \centering
    \includegraphics[width=\columnwidth]{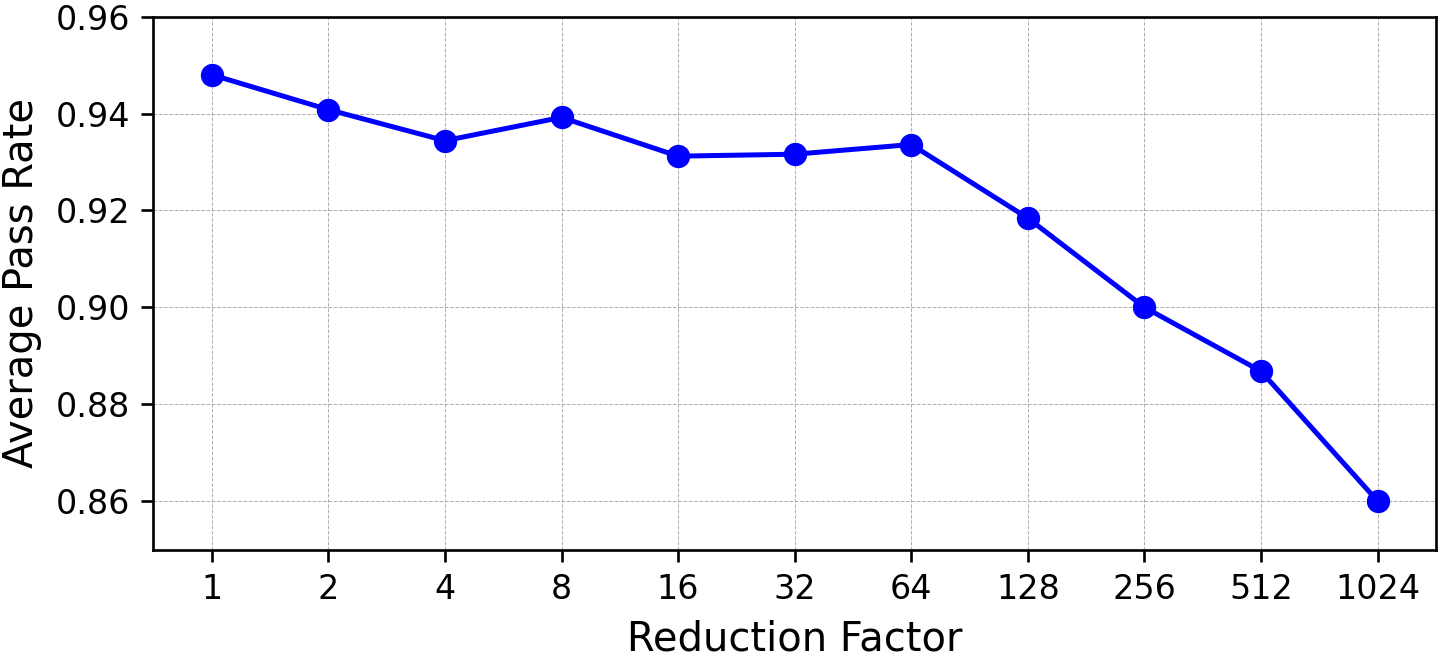}
    \caption{Pass rate trend with event reduction. The system maintains a high detection success rate, even with a reduction factor of 1024$\times$.}
    \label{fig:pass_rate_reduction}
\end{figure}

Figure~\ref{fig:pass_rate_reduction} shows the pass rate as a function of the reduction factor. The pass rate stays above 85\unit{\percent} for reduction factors up to $2^{10}$ (1024), but drops beyond this point, indicating that too much event removal starts to eliminate essential information for localization.

\gls{dvs} are inherently fast due to their asynchronous event-driven nature, allowing near-instantaneous response to changes in the visual field~\cite{Delbruck2008, Posch2014}.  However, we evaluated the potential latency of this system in detecting a press by analyzing event-rate distributions over time, to help understand whether the low-latency of the \gls{dvs}  can be exploited.

\begin{figure}[H]
  \centering
  \includegraphics[width=\textwidth]{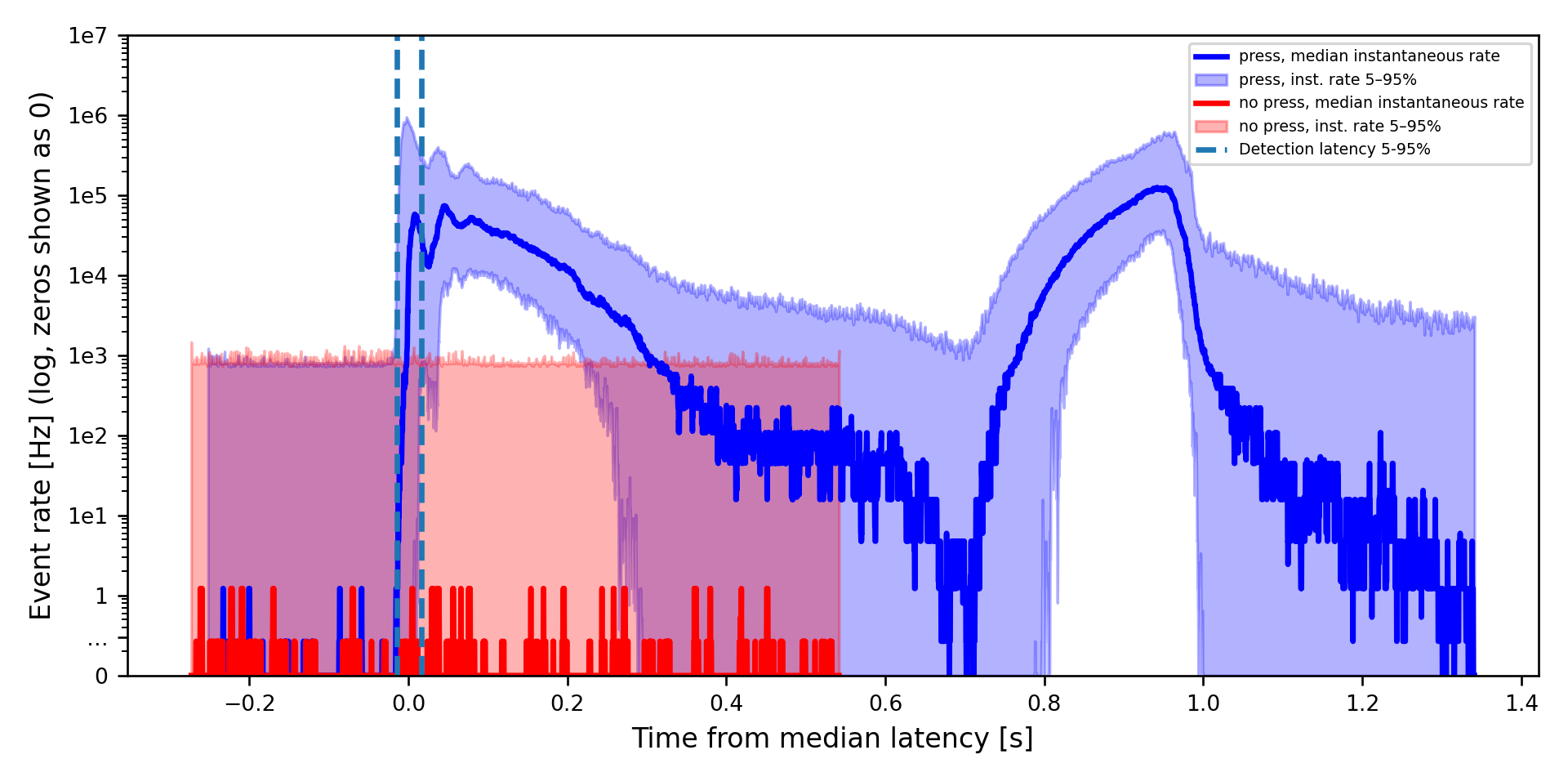}
  \caption{Instantaneous event rates, aggregated over both cameras and all pixels, per press trial (blue) versus background event rate (red), giving median and 5-95\unit{\percent} bounds.  The dashed vertical lines indicate the detection latency distribution of 31~\unit{\milli\second}.}
  \label{fig:event_rate_latency}
\end{figure}

Although the robot triggers were regular at known intervals, no hardware‑recorded ground‑truth onset is available. Therefore, we quantified detection latency distribution by finding the times when the aggregate event rates rise above normal noise and reporting the interval in which most of these crossings occur. For each trial, events were binned at high temporal resolution (0.2~\unit{\milli\second} bins) and smoothed with a short Gaussian kernel ($\sigma = 0.5$~\unit{\milli\second}). The baseline firing rate was estimated from a short background snippet immediately preceding the stimulus. Onset was then defined as the first threshold crossing of a one-sided \gls{cusum} detector tuned to a four-fold rate increase over baseline, requiring at least three consecutive bins above threshold. A threshold parameter \textit{h} was optimized by \gls{roc}
 analysis to ensure $\geq95$\unit{\percent} of trials were detected within a $\pm 100$~\unit{\milli\second} window of the population median onset, while minimizing the false-alarm rate background snippets.\footnote{\url{https://github.com/event-driven-robotics/optoskin/blob/main/latency/latency_from_rates_main.py}} 
Figure \ref{fig:event_rate_latency} shows 95th, 50th, and 5th percentile event rates for press versus no-press, with time aligned to the median onset time during presses (\textit{h}=1).  The detection latency distribution, i.e., the time between 5th and 95th percentile onset times for true positives, is 31~\unit{\milli\second} and the false-alarm rate is 0.13 events/s (where the 31~\unit{\milli\second} is used as a cool-down between consecutive false alarms). 

This window is likely an overestimate of the latency from event production to the possibility of contact detection because the method only looks at event rates combined from both cameras and does not consider any other spatiotemporal cues, which could improve detection. At the same time, it underestimates the end-to-end system latency, since any delay from the robotic trigger to the production of the first press-related events is not included. Furthermore, this method is an analysis of the possibility of contact detection based on event statistics, rather than a proposed online algorithm. All of our processing in this work has been offline and trial-based and we have not presented an online method for press detection, therefore the detection latency distribution is also an underestimate of full-system latency for this reason. With $1024\times$ data reduction, 88\unit{\percent} \gls{tpr} was achieved at h=3, the latency distribution was 113~\unit{\milli\second} with 1.86 false-alarms per second.

\section{Discussion}
In contrast to systems such as~\cite{Chorley2009} and~\cite{Kamiyama2004}, and in common with \gls{msows}~\cite{LoPreti2022}, this system is not dependent on marker tracking. Instead, it observes the effect of surface deformation on the passage of light through the transparent waveguide.

A comparison with \gls{msows} is given in Table~\ref{tab:comparison_dvs_msows}.
\gls{msows} uses a number of infrared emitters (PEs) and detectors (PRs) embedded all around the skin’s periphery (24 of each). 
The PEs are strobed in sequence, the combinations of PEs and PRs define rays across the skin whose precision is limited by emission and reception cones, and the contact is localized by mapping the resulting reductions in received light at each PR to a virtual grid across the skin area (that is, a 16$\times$16 grid with 5~\unit{\milli\meter} cell pitch). 
In contrast, our \gls{dvs}‑based skin uses two 640$\times$480 event‑driven cameras with lenses, in combination with a number of PEs (32) around the skin periphery. 
The PEs are constantly illuminated so that events are not produced in static conditions.
Therefore, the power cost for illumination is elevated in our system, however, synchronization of strobing is not necessary. 
The strobing rate places a limit on the temporal precision of \gls{msows}, which can only be improved by increasing the strobing rate and thus the data rate. 
The \gls{dvs}-based system does not have such a restriction, since event cameras only produce events in the case of change. 
The latency of \gls{dvs} is known to vary with illumination and can be as low as a few microseconds with good illumination.  The experimental scenario and the system presented here do not allow for the potential latency advantage to be demonstrated or evaluated, since a press can be reliably distinguished from background noise after 31~\unit{\milli\second}.
 
Given static illumination, the lenses are necessary to disambiguate the directions from which presses are detected and offer high precision for localization. 
In the method presented here, \gls{dbscan} is used to cluster the events corresponding to a contact. 
The center of the detected cluster is extracted along the horizontal ($u$) axis of each camera and used to determine the angle of incidence based on the camera's field of view, which may, in principle, give an arbitrarily better accuracy. 
\begin{table*}[!t]
  \centering
  \caption{Comparison of this work and MSOWS}
  \label{tab:comparison_dvs_msows}
  \renewcommand{\arraystretch}{1.2}
  \begin{tabular}{p{0.24\linewidth} p{0.24\linewidth} p{0.21\linewidth} p{0.21\linewidth}}
    \toprule
    \textbf{Metric} 
      & \textbf{Lo Preti \emph{et al.} (2022) (MSOWS)} 
      & \multicolumn{2}{c}{\textbf{This Work (DVS-Based)}} \\
    \cmidrule(lr){3-4}
      & 
      & \textbf{No Data Reduction} & \textbf{1024$\times$ Reduction} \\

    \midrule
    Sensor Type              & Optical waveguide (PE-PR) & \multicolumn{2}{c}{DVS (stereo)} \\
    Data Reduction           & N/A                        & No Data Reduction & 1024$\times$ \\
    Number of View Points    & 24                         & 2 &  \\
    Processing Unit          & FPGA DAQ board             & On-board event processing &  \\
    Samples per Press        & 168 samples/PR             & Event stream (sparse) & \\
    ADC Bit Depth            & 12-\unit{\bit}                     & 21-bit & \\
    Press Duration           & 0.144~\unit{\second}                    & 0.55~\unit{\second} & 0.55~\unit{\second} \\
    Bit Rate (press)         & $\sim$145~\unit{\kilo\byte\per\second}             & $\sim$75~\unit{\kilo\byte\per\second} & $\sim$1~\unit{\kilo\byte\per\second} \\
    Bit Rate (idle)          & $\sim$145~\unit{\kilo\byte\per\second}             & $\sim$24~\unit{\kilo\byte\per\second} & $\sim$0.02~\unit{\kilo\byte\per\second} \\
    System Latency           &---                         & 0.031~\unit{\second} Detection latency distribution& 0.113~\unit{\second} Detection latency distribution\\
    Global Error             & 1.8~\unit{\percent}                      & 3.75~\unit{\percent} & $\sim$6.4~\unit{\percent} \\
    RMSE (localization)      &---                         & 4.66 ~\unit{\milli\meter}& 9.33 ~\unit{\milli\meter}\\
    Effective num taxels over sensor area     &---                         & 105 & 34 \\
    Effective num taxels over probed area     &---                         & 46 & 15 \\
    \bottomrule
  \end{tabular}
\end{table*}
The characterization of the localization accuracy is given in terms of the distance between the estimated location and the actual press location, with an \gls{rmse} of 4.66~\unit{\milli\meter}. To contextualize this, an effective number of taxels can be estimated by calculating how many circles of radius 4.66~\unit{\milli\meter} fit into the sensor area. Across the \emph{probed} region of the sensor, this corresponds to 46 effective taxels. (If the same RMSE were to hold across the full sensor area, this would be equivalent to 105 taxels, but we do not claim this, since the RMSE has only been measured within the probed region. It is likely that other effects, such as distortion or the proximity of the LEDs, would be more prominent toward the edges of the sensor.)

For a better comparison with \gls{msows}, the \gls{cmre} statistic for global error in a single press scenario (~\cite{LoPreti2022}) has been computed.
This system achieves a global error of 3.75~\unit{\percent} \textit{versus} \gls{msows}’s 1.8~\unit{\percent}. 

The employed \gls{dvs} cameras are not optimized for this application and may be highly overspecified. 
It may be that small pods of \gls{dvs} pixels in 2 or more locations could deliver more efficient performance. 
To evaluate the possibility of using fewer event data, events have been randomly eliminated within trials, and the results shown in the table~\ref{tab:comparison_dvs_msows} are for a 1024$\times$ reduction. 
In this case, the \gls{cmre} rises to 6.4~\unit{\percent}, \gls{rmse} rises to 9.33~\unit{\milli\meter}, the effective number of taxels reduces to 34~(15 probed), and pressure detection latency distribution is approximately 31~\unit{\milli\second} and remains at about 113~\unit{\milli\second} under a 1024$\times$ data reduction.
By applying a 1024$\times$ data reduction, the \gls{dvs} version can cut its peak press‑mode bit‑rate from 75~\unit{\kilo\byte\per\second} down to 1~\unit{\kilo\byte\per\second}, and its idle bit‑rate from 24~\unit{\kilo\byte\per\second} to 0.02~\unit{\kilo\byte\per\second}, compared to \gls{msows}’s constant 145~\unit{\kilo\byte\per\second}.
\gls{msows} used 12-bit A-D conversion, whereas we have considered 21 bits per \gls{dvs} event, given the size of the VGA address space.

In addition to comparing our system with \gls{msows}, we place it alongside other recent tactile localization technologies. These include resistive and tomographic skins, capacitive arrays, optical tactile sensors, magnetic skins, inductive and microwave sensors, and acoustic approaches. These systems vary widely in materials, wiring complexity, sensing principles, and achievable accuracy, but all aim to localize contact across a surface. The main differences between our \gls{dvs}-based skin and these technologies concern sensing modality and data rate: our approach uses two event cameras instead of dense arrays, avoids global inverse problem solving, and naturally supports low idle bit-rate operation. A summary comparison of these technologies is provided in Table~\ref{tab:comparison_tactile_sensors}.

\begin{table*}[t]
\centering
\caption{Comparison of representative tactile localization technologies.}
\label{tab:comparison_tactile_sensors}
\renewcommand{\arraystretch}{1.25}
\begin{tabular}{p{4.8cm} p{3.0cm} p{2.6cm} p{2.6cm} p{3cm}}
\hline
\textbf{Sensor} &
\textbf{Localization Capabilities} &
\textbf{Materials} &
\textbf{Accuracy / Resolution} &
\textbf{Techniques (Hardware/Algorithm)} \\
\hline

 \gls{eit} sheet (\cite{Chen_etal25_EIT}) &
Multi-touch with bending &
Conductive hydrogel; elastomer layer &
5.4~\unit{\milli\metre} error &
\gls{eit} reconstruction with ML for touch–bend decoupling \\

\addlinespace

Hydrogel \gls{eit} skin (\cite{HARDMAN2023}) & Multi-touch & self-healing ionic hydrogel & 12.1~\unit{\milli\metre} & Data-driven \gls{eit} reconstruction \\

\addlinespace

Stretchable capacitive array (\cite{Sarwar_etal17_StretchCapacitive}) &
Multi-touch under stretch/bending &
Transparent elastomer; stretchable electrodes &
1~\unit{\centi\meter} resolution &
Capacitive grid with deformable substrate \\

\addlinespace

Magnetic skin (\cite{Hu2024}) &
Multi-touch, multi-scale; large-area sensing &
Magnetized elastomer film; Hall sensors &
1.2~\unit{\milli\metre} error over 48400~\unit{\milli\metre} area &
signal proceesing withh \gls{cnn} \\

\addlinespace

Graphene Hall magnetic array (\cite{Li2025}) &
Single-touch with virtual 6×6 resolution &
Graphene Hall elements; magnetized layer &
1.3~\unit{\milli\metre} average error &
Vertical periodic magnetization to reduce field interference \\

\addlinespace

Flexible EIT tactile skin (\cite{Chen2025}) &
Multi-touch with bending &
Magnetic hydrogel layer and Ecoflex  &
4.8 $\pm$ 2.8~\unit{\milli\metre} touch error; bend RMSE $\sim$0.9$^\circ$ &
\gls{cnn} state classifier\\

\addlinespace

SonicBoom (\cite{Lee2025}) &
Single-touch &
MEMS microphones &
0.43--2.22~\unit{\centi\meter} localization error &
Vibration-based triangulation with learned model \\

\hline
\end{tabular}
\end{table*}

Only single-touch localization has been investigated in this work. Low-pressure detection and multi-touch interactions were not studied. Early tests (not shown here) agree with previous findings in \gls{msows} that contact can still be detected through internal light reflection when the surface is bent. With the reduced viewing locations provided by focused optics, the present system could also be extended to larger or more complex surfaces. Beyond localization, we did not attempt any further tasks such as identifying the type of contact or classifying the source of the pressure. However, the event data contain richer information than just position. Features such as the shape of the event cluster, the number of events produced, and the timing of the event activity could support future work on force estimation, slip detection, object recognition, or classification of different touch types. These tasks would require additional processing and calibration and remain outside the scope of this proof-of-concept study, but they represent clear next steps.

\subsection{Limitations and Future Directions for Multi-Touch Sensing}

A key limitation of the current work is its exclusive focus on single-point contact localization. The methodology presented, which relies on identifying a single dominant event cluster in each camera, is fundamentally unequipped to resolve multiple simultaneous touch events. This is a critical consideration for future applications in complex, interactive scenarios.

Another limitation is that the current system does not estimate the magnitude of the applied pressure. This work was designed as a proof of concept focused on press localization. The robot pressed the sensor surface with a cubic tip to a fixed indentation depth of 2~\unit{\milli\meter}; position control was used and the applied force was not measured during the experiments. However, the event data suggest that pressure estimation could be added in future work. As shown in Fig.~\ref{triangulation} C--D, a single press produces a characteristic cluster of events on each camera. By analyzing features such as the size of the activated region, the density of events, or the temporal evolution of the cluster during indentation, it may be possible to infer the applied force. A dedicated calibration procedure would be required to map these features to physical pressure values.

Another limitation of the present work is that the evaluation was carried out only within the central region of the sensor surface. In early tests, presses near the edges of the silicone produced noticeably larger localization errors, as the image distortion increases toward the periphery. More importantly, the silicone layer had to be made very soft (30:1 PDMS, see Section~\ref{design}) to maximize sensitivity, and this made the material mechanically fragile at the edges. Since the NIR LEDs are embedded directly in the silicone along the boundaries, pressing too close to an edge caused the silicone to lift and risk damage to the LEDs or tearing of the material. For these reasons, the robot was constrained to press only within the safer central region of the sensor during the experiments, corresponding to 52.7\unit{\percent} of the usable sensing area. A more robust fabrication process, would be needed to allow full-area testing in future work.

The failure arises from the classic stereo correspondence problem. In a multi-touch scenario, each \gls{dvs} camera would detect multiple, distinct clusters of events. The current algorithm would either erroneously merge nearby clusters into a single, inaccurate centroid or, if modified to detect multiple centroids, would lack a mechanism to correctly pair the corresponding centroids from the left and right cameras. For instance, with two presses generating centroids $L_1$ and $L_2$ in the left camera's view and $R_1$ and $R_2$ in the right's, the system could not disambiguate between the correct pairings ($L_1$-$R_1$, $L_2$-$R_2$) and the incorrect 'ghost' pairings ($L_1$-$R_2$, $L_2$-$R_1$), leading to erroneous localization.

However, the event-based nature of the DVS offers a powerful solution. It is highly improbable that two distinct mechanical presses would be initiated in the exact same microsecond. Future work can implement a temporal correlation filter. By segmenting event clusters based on their onset timestamps, the system can pair clusters from the two cameras that appear and evolve within a narrow, shared time window (e.g., $\leq 1$ ms), thus robustly solving the correspondence problem for non-synchronous contacts.

An alternative approach is to move beyond simple centroiding and utilize richer spatial features of the event clusters. The shape, size, total event count, and velocity profile of corresponding clusters should be similar across both camera views. A future algorithm could combine epipolar geometry constraints with a feature-matching technique (e.g., comparing cluster moments or event density histograms) to calculate a matching score and identify the most probable pairings.

For guaranteed disambiguation in cluttered or highly dynamic scenarios, the system could be extended with a third, non-collinear camera. This would provide an additional geometric constraint, allowing for robust triangulation even with ambiguous pairings between any two of the views. The intersection point validated by all three viewpoints would confirm a true contact, while 'ghost' points would be rejected for failing this multi-view consistency check.

Beyond the current triangulation-based method, an alternative approach to resolving the stereo correspondence problem in \gls{dvs}-based tactile sensing is the use of \glspl{snn}. \cite{Osswald2017} proposed an \gls{snn} model that enables 3D perception by leveraging spike-based representations and temporal dynamics.

\section{Conclusions}
This work presents an optical skin system designed for advanced tactile sensing using \gls{dvs} cameras with a compliant silicone layer. 
Based on the \gls{msows} design~\cite{LoPreti2022}, this work proposes a marker-free event-based tactile sensor by replacing embedded photoreceivers with \glspl{dvs} cameras. 
Our approach allows for localization of contacts through stereo vision and triangulation, with a simple, flexible, and responsive design suitable for dynamic tactile applications.
The system achieves a pressure localization accuracy of 4.66~\unit{\milli\meter}~\gls{rmse}. 
Data reduction analysis reveals that the event-based approach could sustain a pass rate above 85~\unit{\percent} even when reducing data transmission by a factor of 1024, confirming the feasibility of a low-bandwidth implementation. 
The detection latency distribution is  31~\unit{\milli\second}, demonstrating the potential system's suitability for real-time tactile interactions.

By utilizing \gls{dvs} technology, the system benefits from high temporal resolution and reduced data bandwidth, making it well-suited for scenarios requiring real-time processing and high-speed interaction. 
Furthermore, the system’s marker-free design simplifies integration, paving the way for scalable and adaptable soft robotic applications.

\section{Future Work}

While the current design is a proof-of-concept that successfully localizes contacts, several areas for improvement remain.

Future work will focus on a new design that grants the optical skin higher sensitivity. 
The next envisioned steps include extensive testing on various tactile tasks, such as precise localization, slip detection, and force mapping. 
These would be useful in fine-tuning the system to achieve better accuracy and reliability, and become more easily deployable in soft robots and interactive objects. 
Unused data features such as V-address, polarity, and precise timing may be used to increase the system's discrimination ability. 
Optics may be better matched to the employed cameras.
Smaller cameras with fewer pixels may provide acceptable performance with even lower data rates. 
Moving towards a design with small clusters of pixels and dedicated optics introduces key trade-offs. 
While it reduces the overall data processing load and lowers bandwidth requirements, it increases wiring complexity. 
Additional synchronization strategies may be required between clusters to maintain localization accuracy. 
Furthermore, decreasing the number of pixels limits the resolution of event detection, potentially impacting fine-grained spatial accuracy. 
However, a well-optimized optical system could compensate for these constraints, enabling a balance between data efficiency and sensing performance.
\gls{snn}-based methods might be investigated as an alternative to triangulation, potentially improving accuracy and resolving remaining systematic skew and adaptability to more complex tactile interactions.  

\section*{Acknowledgments}
CB, SB, MK, PT and LB acknowledge the support of the IIT Flagship program Brain\&Machines and Technologies for Sustainability.
CB acknowledges the financial support of the National Biodiversity Future Center funded under the National Recovery and Resilience Plan (NRRP), Mission 4 Component 2 Investment 1.4 - Call for tender No. 3138 of 16 December 2021, rectified by Decree n.3175 of 18 December 2021  of Italian Ministry of University and Research funded by the European Union – NextGenerationEU. SB acknowledges the financial support from PNRR MUR Project PE000013 "Future Artificial Intelligence Research (hereafter FAIR)", funded by the European Union – NextGenerationEU.
SB acknowledges the support of the European Union Horizon Europe programme, grant agreement ID: 101120727, "PRIMI: Performance in Robots Interaction via Mental Imagery"

\bibliographystyle{plainnat}
\bibliography{bibliography}

\end{document}